\documentclass[11pt, letterpaper]{article}
\usepackage[margin=1in]{geometry}
\usepackage[utf8]{inputenc} 

\usepackage[sort,compress]{natbib}
\setcitestyle{numbers}

\usepackage[format=plain,labelfont={bf},textfont=it]{caption}
\usepackage{graphicx}

\usepackage{subcaption}
\DeclareCaptionLabelFormat{ms_label_style}{\sffamily\itshape\small\linespread{1}\selectfont#2}
\DeclareCaptionFont{ms_font_style}{\sffamily\itshape\small\linespread{1}\selectfont}
\usepackage[table,xcdraw]{xcolor}
\usepackage{comment}

\usepackage{totcount}
\newtotcounter{citnum} %
\def\oldbibitem{} \let\oldbibitem=\bibitem
\def\bibitem{\stepcounter{citnum}\oldbibitem}

\title{Designing Equitable Algorithms}

\author{
  Alex Chohlas-Wood\\
  Stanford University\\
  \texttt{alexcw@stanford.edu}
  \and
  Madison Coots\\
  Harvard University\\
  \texttt{mcoots@g.harvard.edu}
  \and
  Sharad Goel\\
  Harvard University\\
  \texttt{sgoel@hks.harvard.edu}
  \and
  Julian Nyarko\\
  Stanford University\\
  \texttt{jnyarko@law.stanford.edu}
}

\date{}

\begin{document}
\maketitle

\thispagestyle{empty}

\begin{abstract}
Predictive algorithms are now used to help distribute a large share of our society's resources and sanctions, such as healthcare, loans, criminal detentions, and tax audits.
Under the right circumstances, these algorithms can improve the efficiency and equity of decision-making.
At the same time, there is a danger that the algorithms themselves could entrench and exacerbate disparities, particularly along racial, ethnic, and gender lines.
To help ensure their fairness, many researchers suggest that algorithms be subject to at least one of three constraints: 
(1) no use of legally protected features, such as race, ethnicity, and gender;
(2) equal rates of ``positive'' decisions across groups; and
(3) equal error rates across groups.
Here we show that these constraints, while intuitively appealing, often worsen outcomes for individuals in 
marginalized groups, and can even leave all groups worse off.
The inherent trade-off we identify between formal fairness constraints and welfare improvements---particularly for the marginalized---highlights the need for a more robust discussion on what it means for an algorithm to be ``fair''. We illustrate these ideas with examples from healthcare and the criminal-legal system, and make several proposals to help practitioners design more equitable algorithms. 
\end{abstract}

\section{Introduction}
\label{intro}

With the advancement of statistical methods, computational resources, and data availability, the last decade has seen a dramatic increase in the use of algorithmic decision-making across all facets of life.
Banks make algorithmic predictions to assess who is at risk of default and should thus not be offered a loan~\citep{leo2019machine},
and to identify possible instances of money laundering~\citep{zhang2019machine}.
In healthcare, 
algorithms are used to decide who gets screened for diseases
like diabetes~\citep{agarwal2022diabetes},
and to allocate resource-limited benefits such as kidney transplants~\citep{friedewald2013kidney} and HIV-prevention counseling~\citep{wilder_2021}.
Criminal justice agencies use algorithms to inform the allocation of police resources~\citep{mohler2015randomized,doucette2021impact},
to assist investigators~\citep{chohlas2019recommendation,o2019facial},
to inform incarceration and sentencing decisions~\citep{demichele2018public,goel2021accuracy,skeem2016gender},
and to limit the impact of perceived race on prosecutorial charging decisions~\citep{chohlas2021blind}.
Technology companies use algorithms to decide who sees ads for housing~\citep{pmlr-v81-speicher18a}
and employment opportunities~\citep{lambrecht2019algorithmic}, among others.
Child services agencies use algorithms to estimate the risk of an adverse event like child abuse~\citep{de2020case,chouldechova2018case,brown2019toward,shroff2017predictive}.
City agencies use algorithms to prioritize building inspections~\citep{mayer2013big}.
School districts use algorithms to assign students to their preferred school~\citep{allman2022designing} 
and to identify students who are at risk of falling behind on learning material~\citep{cattell2021identifying}.

While it appears that the use of algorithms for critical decision-making will only increase in the near future, 
some have pointed to a heightened danger that these same algorithms could influence decision-making in a way that is unfair to marginalized groups, such as racial or ethnic minorities~\citep{o2016weapons,eubanks2018automating}. Legal scholars have argued that certain forms of algorithmic decision-making may even be in conflict with important constitutional or regulatory protections granted to groups defined by race and ethnicity, rendering them impermissible~\citep{huq2018racial,yang2020equal,hellman2020measuring,mayson2019bias,barocas2016big}.
In response to these concerns, researchers have developed several fairness criteria with the goal of ensuring that algorithms achieve equitable decision-making~\citep{corbett2023measure,mitchell2021algorithmic,chouldechova2017fair}.
These criteria range from excluding certain legally protected characteristics---such as race, ethnicity, gender, and their close correlates---from an algorithm's inputs,
to requiring certain key metrics, like error rates, be equal across demographic groups.

Today, adherence to these fairness constraints has become common practice in the design of algorithms across many contexts.
However, a dogmatic implementation of these constraints often comes at the cost of inflicting additional burdens on individuals in all groups, including those in marginalized communities.
For instance, in medicine, common diabetes risk calculators that ignore a patient's race and ethnicity systematically underestimate diabetes risk for Asian, Hispanic, and Black patients and overestimate diabetes risk for White patients~\cite{agarwal2022diabetes}. 
There may well be good reasons to exclude the use of race in medical diagnoses---e.g., to guard against pernicious attitudes of biological determinism---but this constraint comes at the cost of 
poorer treatment for patients in every group.
In Section~\ref{sec:consequences} we lay out this tension between fairness constraints and welfare in more detail. Then, in Section~\ref{sec:thepath}, we make several recommendations to address this tension between fair processes and fair outcomes, as well as other problems commonly encountered when building algorithms.
We hope our discussion helps researchers, policymakers, and practitioners understand the subtleties of popular fairness constraints, and leads to the design of more equitable algorithms.

\section{Popular fairness constraints and their consequences}
\label{sec:consequences}

Over the last several years,  
researchers across numerous fields 
have considered the equitable design of algorithms, 
including in computer science and statistics~\citep{chouldechova2017fair, hardt2016equality, buolamwini2018gender,kleinberg2016inherent,corbett2017algorithmic,dwork2012fairness,chouldechova2020snapshot,coston2020counterfactual,loftus2018causal,zafar2017parity,zafar2017fairness,woodworth2017learning,wang2019equal,carey2022causal,kusner2017counterfactual, nabi2018fair, wu2019pc,galhotra2022causal,mhasawade2021causal, kilbertus2017avoiding,chiappa2019path,zhang2018fairness,zhang2016causal,L2BF,nilforoshan2022causal},
law~\citep{mayson2019bias,huq2018racial,yang2020equal,bent2019algorithmic,chander2016racist,kim2022race,ho2020affirmative},
medicine~\citep{mccradden2020ethical,paulus2020predictably,goodman2018machine,pfohl2021empirical,obermeyer2019dissecting},
the social sciences~\citep{imai2020principal,imai2020experimental,berk2021fairness,kleinberg2018discrimination,cowgill2019economics,nyarko2021breaking,grgichlaca2022,liang2021algorithmic},
and philosophy~\citep{card2020consequentialism,hu2020s,kasy_2021}.
Many of these studies have proposed formal statistical principles for designing ``fair'' algorithms.
Here we group these myriad fairness principles into three conceptual categories: 
\begin{enumerate}
    \item \emph{Blinding}, in which one limits the effects of demographic attributes---like race---on decisions;
    \item \emph{Equalizing decision rates} across demographic groups; and
    \item \emph{Equalizing error rates} across demographic groups.
\end{enumerate}

To many, these principles represent intuitively appealing understandings of fairness, and they have been applied to a variety of contexts in which algorithms guide decisions.
They are often implemented with the explicit goal of protecting members of disadvantaged communities, 
but, as we discuss next, strict adherence to these principles often leads to worse outcomes for those in marginalized groups---and society as a whole~\citep{corbett2023measure}.

To illustrate with a practical example,
consider the case of diabetes risk estimation.
Approximately one in ten Americans suffer from Type 2 diabetes, which can lead to other serious health problems, including heart disease, kidney disease, and vision loss. 
Upon learning of their diagnosis, patients can better manage their condition---for example, through changes in diet and physical activity---making early detection critical to improving health outcomes. 
In theory, every patient could be screened at regular intervals in an effort to detect diabetes early. 
But screening itself comes with monetary and non-monetary costs 
(e.g., patients may need to take time off from work, resulting in lost income).
The medical community accordingly recommends that only those with at least a moderate risk of developing diabetes undergo screening. 
For example, results by \citet{agarwal2022diabetes} suggest that patients will typically benefit from screening if their risk of diabetes is above 1.5\%.
To follow this recommendation,
statistical risk algorithms can be used to estimate the diabetes risk for every patient,
offering screening to those with 
an estimated risk above 1.5\%.

We empirically ground our discussion by training statistical models that estimate diabetes risk using data from the National Health and Nutrition Examination Survey (NHANES).
NHANES combines interview responses with laboratory data to provide insight into the health and nutritional status of adults and children in the U.S. 
The survey is conducted every two years by the National Center for Health Statistics
and is frequently used by researchers to assess the prevalence of major diseases and their risk factors across the U.S.\ population. 
In our analysis, we use the four NHANES cycles from 2011--2018. 
Following \citet{agarwal2022diabetes}, we restricted our sample to 18,000 patients who were not pregnant, 
were 18--70 years old, 
and had a BMI between 18.5 kg/m$^2$ and 50.0 kg/m$^2$.

We now discuss the three fairness constraints above, in turn showing 
how statistical risk algorithms that adhere to each %
constraint
may lead to worse outcomes for minority and majority groups alike.

\subsection*{The consequences of blinding}
In an attempt to limit the effects of demographic attributes on risk assessments, the principle of blinding mandates that algorithms not have access to certain demographic characteristics, such as race or ethnicity, when estimating patient risk.%
\footnote{A related family of causal fairness criteria seeks to reduce both the direct and indirect effects of race on decisions~\citep{carey2022causal}, since even if algorithms are formally ``blind'' to race, race may still impact decisions indirectly through the algorithm's other inputs.
This line of work, however, suffers from at least two serious limitations.
First, it requires formalizing a causal effect of race, a long-standing statistical and conceptual problem replete with  challenges that are succinctly captured by the mantra, ``no causation without manipulation''~\citep{holland1986statistics}. In particular, one must make sense of counterfactuals in which a person's race is altered~\citep{gaebler2020causal,hu2020s,greiner2011causal,sen2016race},
a notion that is difficult, and perhaps impossible, to make precise.
Second, recent mathematical results have shown that these causal fairness definitions constrain algorithms so severely that they often produce unintended results~\citep{nilforoshan2022causal,corbett2023measure}.
For example, under one prominent causal fairness definition, the only permissible algorithm in many situations is one that makes the same decision for every individual, irrespective of their risk factors.
}
For example, diabetes risk may be estimated by a statistical risk algorithm that considers one's age and body mass index (BMI), but not their race or ethnicity.
This principle is also sometimes called
``fairness through unawareness.''

\begin{figure}[t]
	\centering
	\begin{subfigure}{.46\columnwidth}
		\includegraphics[width=\columnwidth]{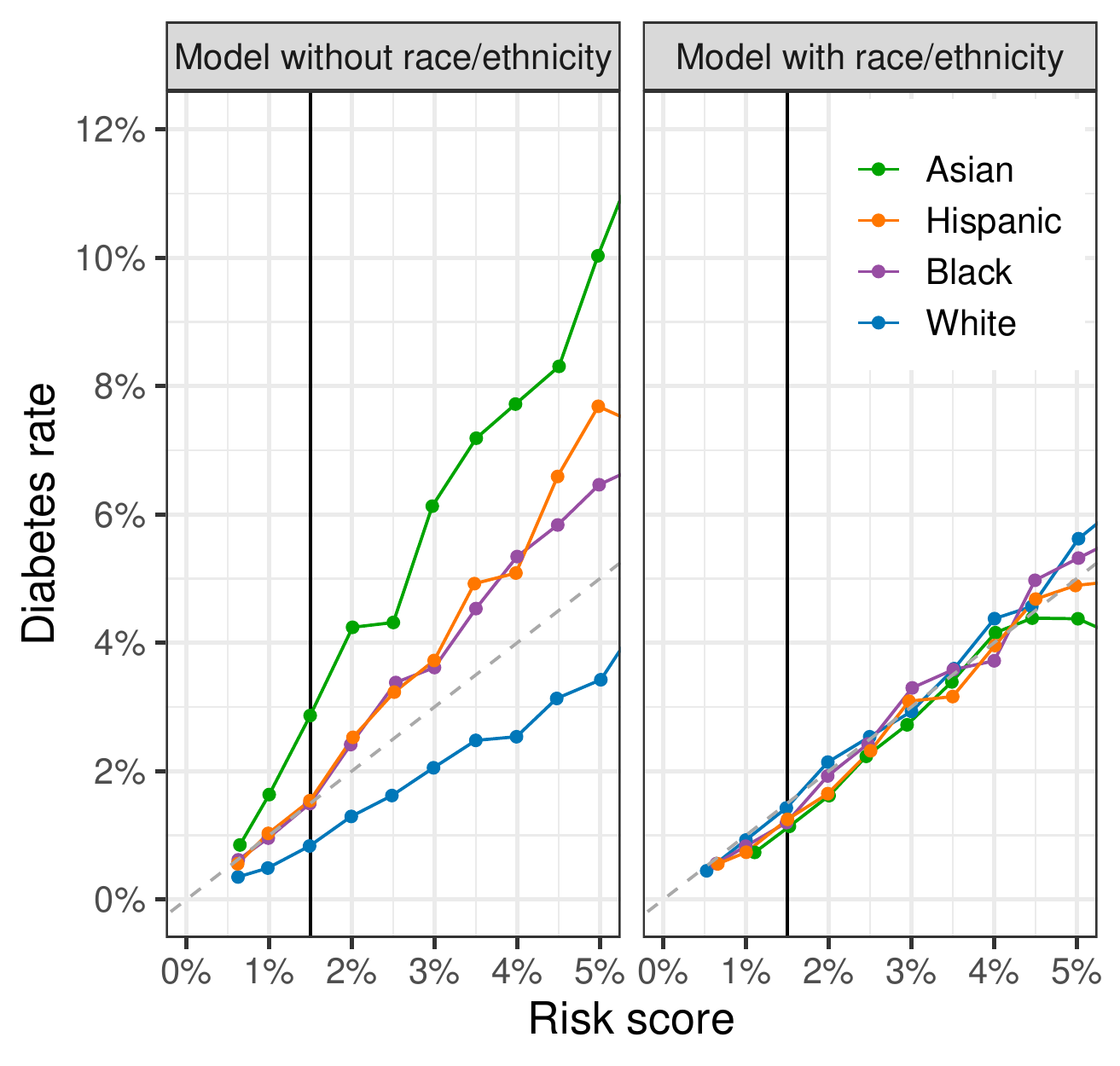}
		\caption{Estimated vs.\ actual diabetes risk under models that exclude (left) or use (right) race and ethnicity.}
		\label{fig:calibration-facet}
	\end{subfigure}
    \hspace{5mm}
    \begin{subfigure}{.46\columnwidth}
        \includegraphics[width=\columnwidth]{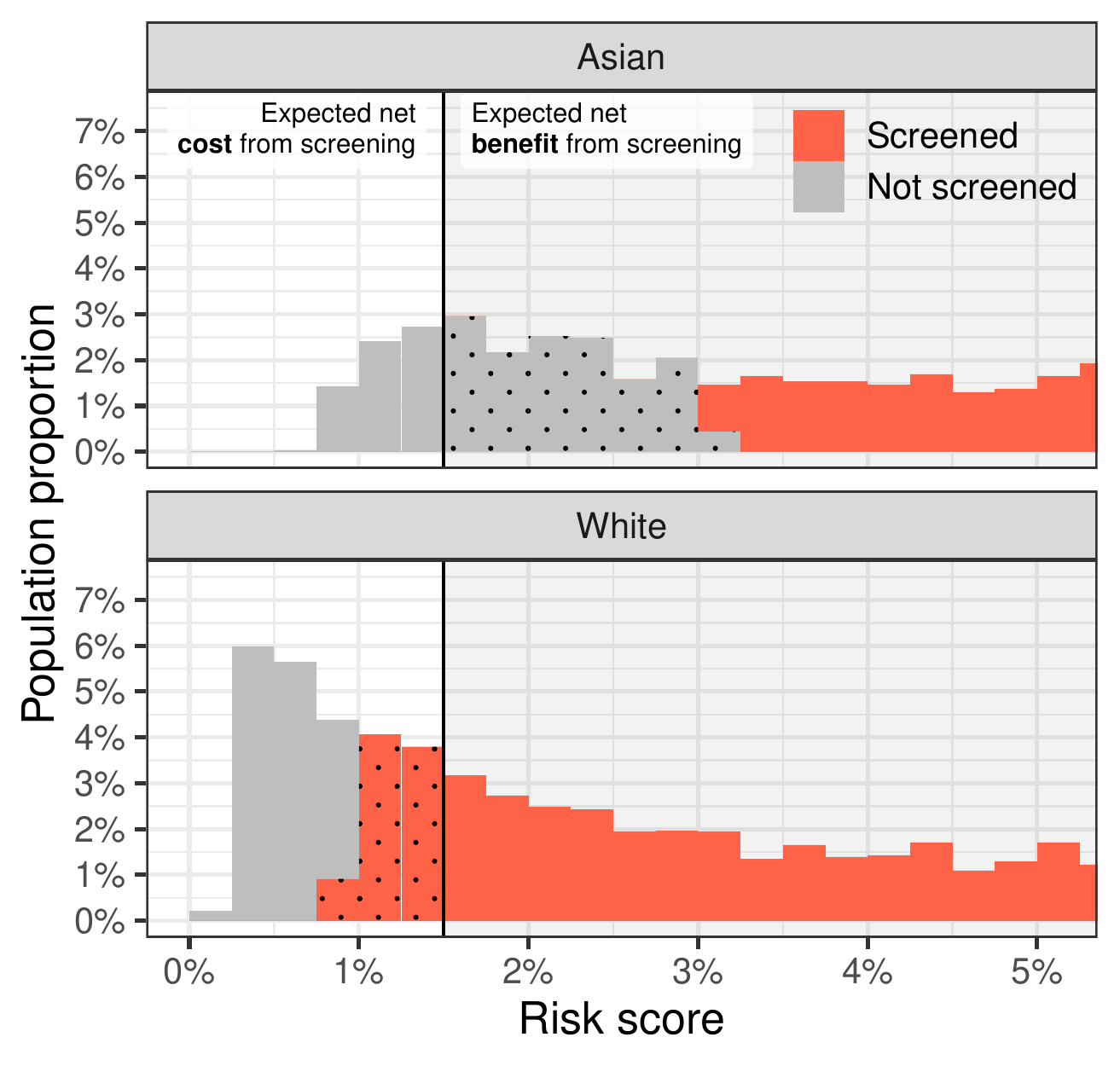}
        \caption{The relative cost of excluding race and ethnicity from diabetes risk estimates.}
        \label{fig:histogram}
    \end{subfigure}
 
	\vskip 0.1in
	\caption{The plots in \subref{fig:calibration-facet} compare the estimated risk from 
             race-blind and race-aware models against the observed rates of diabetes across demographic groups.
             The plots in \subref{fig:histogram} illustrate the cost of using a race-blind model
             for Asian and White patients
             when compared against (more accurate) race-aware risk scores. The plot shows the distribution of race-aware risk scores for Asian and White patients, and the shaded areas show which patients would receive screening under the race-blind model.
             The dotted area, in particular,
             covers patients for whom the race-blind model makes a screening error,
             either because it fails to recommend Asian patients for screening 
             (even though they would expect to benefit from a test)
             or because it recommends White patients for screening 
             (when they would not expect to benefit from a test).
             }
	\label{fig:calibration}
\end{figure}

In Figure~\ref{fig:calibration-facet}, 
we compare diabetes risk estimated by
 models that either exclude (left panel) or include (right panel) information on a patient's race and ethnicity
against empirical rates of diabetes prevalence.
The model that is blind to a patient's race and ethnicity systematically underestimates diabetes risk for Asian, Black, and Hispanic patients, 
while it systematically overestimates diabetes risk for White patients---a problem that does not occur in the model that considers race and ethnicity.

Under the blind model, the ``miscalibrated'' risk scores could in turn
result in erroneous screening decisions
for some patients.
Imagine, for concreteness, a hypothetical 
30-year-old Asian patient with a BMI of 21.5 kg/$m^2$.
Under the blind model, our hypothetical patient would have an estimated diabetes risk of 
approximately 1.15\%, and so would not be screened
based on the 1.5\% screening threshold.
However, as shown in the left panel of 
Figure~\ref{fig:calibration-facet},
Asian patients with a nominal, race-blind diabetes risk of 1.15\% 
have an empirical rate of diabetes close to 2\%.
We accordingly expect our hypothetical patient to benefit from screening,
even though the race-blind model would recommend against it.
The race-aware model, in contrast, correctly estimates that patients like this have an elevated risk of diabetes and thus recommends they be screened.

Analogously, consider a hypothetical 40-year-old White patient with a BMI of 20.5 kg/$m^2$. 
The race-blind model estimates our hypothetical patient has a 1.9\% risk of diabetes, but, in reality, only 1.3\% of patients like this  have diabetes. 
The race-blind model would recommend our patient be screened, even though we expect screening to impose a net cost in this case.
As before, the race-aware model correctly estimates that patients like this have a relatively low risk of diabetes and thus advises against screening.

Generally,
if diabetes risk is estimated without the use of race and ethnicity,
some non-White patients expected to benefit from screening
would be counseled against screening,
while some White patients expected to incur a net cost from screening 
would be screened anyway.
On the other hand, models that use a patient's race and ethnicity are able to account for differences in diabetes risk across groups, and so do not make these systematic errors.

In Figure~\ref{fig:histogram},
we show the overall consequence of banning race and ethnicity from the algorithmic inputs
on the two groups with the largest disparity, Asian and White patients.
When avoiding the use of race and ethnicity,
nearly 14\% of Asian patients would not receive a screening even though they would be expected to benefit from it.
Similarly,
about 9\% of White patients would be screened for diabetes
even though they would be expected to incur a net cost from screening.

\subsection*{The consequences of equalizing decision rates}

\begin{figure}[t]
	\centering
	\includegraphics[width=0.8\columnwidth]{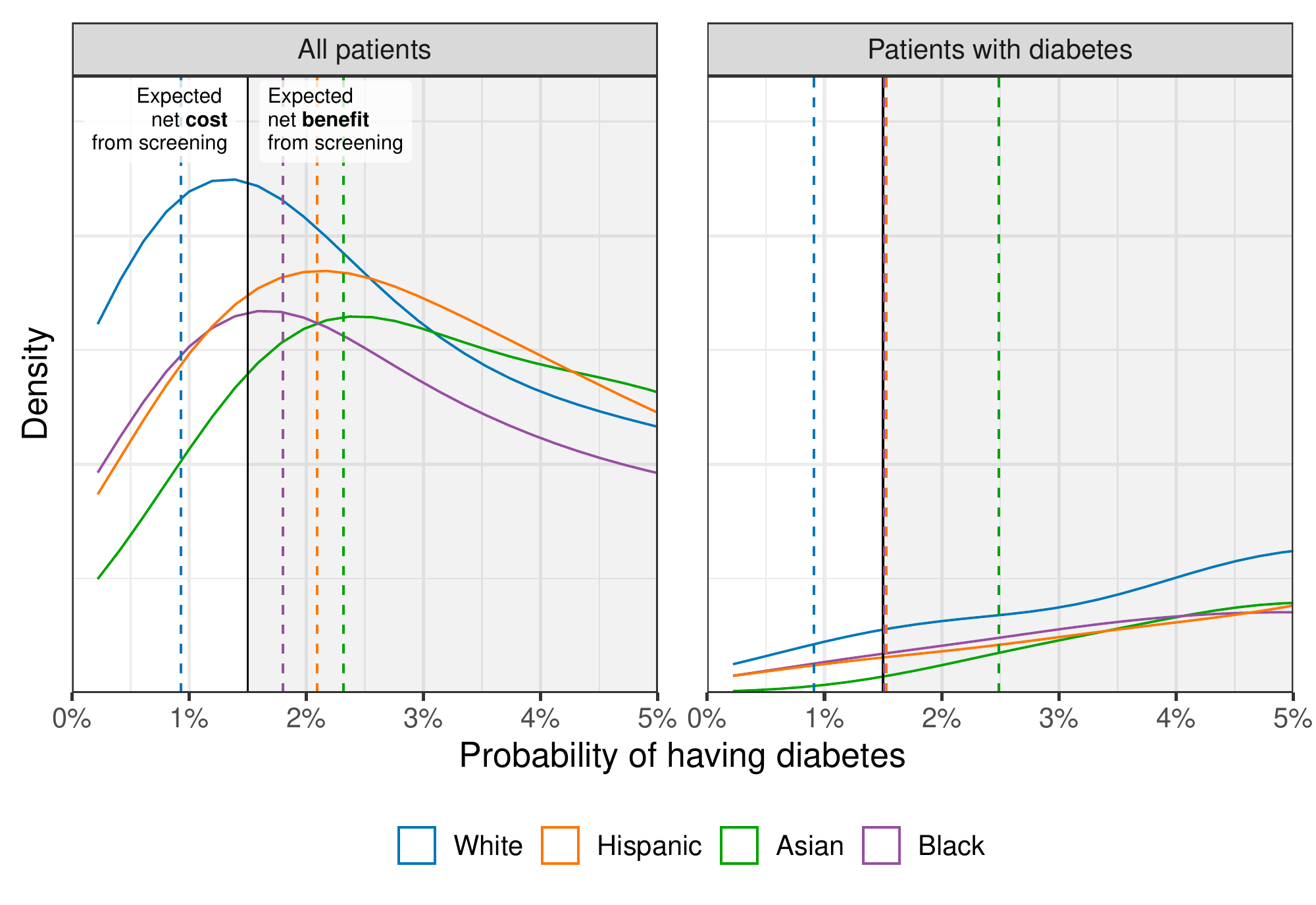}
	\vskip 0.1in
	\caption{The distribution of diabetes risk for all patients (left) and patients with diabetes (right).  Estimates of risk were generated using patients' age, BMI, and race or ethnicity. The dashed vertical lines correspond to screening thresholds that equalize decision rates across groups (left) and false negative rates across groups (right).}
	\label{fig:risk-distributions}
\end{figure}

A second common fairness constraint requires that algorithmic decisions be made at equal rates across demographic groups---defined, for example, by race and ethnicity. 
For instance, a policy following this  constraint 
might enforce that the proportion of White patients 
who are recommended for diabetes screening is approximately equal to the proportion of Asian patients  recommended for screening.
As with blinding, equalizing decision rates may feel intuitively appealing. 
But---also as with blinding---equalizing decision rates can likewise impose considerable costs on members of every group, due to
the ``problem of infra-marginality''~\citep{simoiu2017problem,ayres2002outcome,galster1993,carr1993,knowles2001,engel2008,anwar2006,pierson2018fast}.

Returning to our running diabetes example, 
consider risk scores that are based on a patient's
age, BMI, and race or ethnicity.
(Similar issues result if we start with the blind risk scores.)
The left panel of Figure~\ref{fig:risk-distributions} shows the 
distribution of estimated risk scores, disaggregated by race and ethncity.
In this case, 
76\% of White patients have risk scores above the 1.5\% screening threshold (indicated by the vertical black line), but 93\% of Asian patients, 90\% of Hispanic patients,
and 88\% of Black patients are above the threshold.
As a result, 
if we make the optimal decision for each individual patient---screening them if their likelihood of having diabetes is above 1.5\%---we would violate the principle of equalizing decision rates.

To equalize screening rates across racial and ethnic groups,
we could set group-specific screening thresholds.
Under a single, non-group-specific screening threshold of 1.5\%,
approximately 85\% of individuals are screened.
To equalize screening rates,
we could similarly choose to screen the riskiest 85\% of each group.
The vertical lines in the left panel of Figure~\ref{fig:risk-distributions} show the corresponding group-specific screening thresholds for this policy.
Under this approach,
we would screen White patients with a risk score of approximately 1\% or above, which includes many relatively low risk White patients---namely those with risk between 1\% and 1.5\%---for whom we expect screening to impose net costs.
Conversely, we would only screen Asian patients who have relatively high risk of diabetes, above approximately 2.5\%.
In this case, we would fail to screen many Asian patients for whom we expect screening to have net benefits.
By equalizing decisions rates, we thus harm members of all racial and ethnic groups.

\subsection*{The consequences of equalizing error rates}
A third popular class of fairness constraints requires that error rates be equal across groups.
In the context of our running example, one might, for example, demand that the false negative rate 
of screening decisions be the same across racial and ethnic groups. This constraint means that among those who in reality have diabetes, the proportion who are not screened is the same across groups.
As with blinding and equalizing decision rates, equalizing error rates has intuitive appeal, but, as with those other constraints, it can harm members of all groups.

In right panel of 
Figure~\ref{fig:risk-distributions}, 
we show the distribution of estimated risk among those 
patients who have diabetes, disaggregated by race and ethnicity.
Under a policy of screening patients above a 1.5\% threshold (indicated by the vertical black line),
the false negative rate for a group corresponds to the area under that group's density curve that is to the left of the threshold.
Specifically, White patients have a 1.7\% false negative rate, meaning they would not be recommended for screening
even though they have diabetes.
In comparison, the false negative rate for Asian patients is less than 0.1\%. 
As above, making the optimal screening decision for each patient would violate the principle of equalizing error rates.

Also as above, we could equalize false negative rates
by setting group-specific screening thresholds.
For example, if we screen White patients above a 0.9\% threshold, and Asian patients above a 2.5\% threshold, then both groups would have a false negative rate of 0.7\%.
But such a policy would mean that we recommend screening for some relatively low-risk White patients and do not recommend screening for some relatively high-risk Asian patients, harming some individuals in both groups.

\subsection*{Trade-offs in resource-constrained settings}
\begin{figure}[t]
	\centering
	\begin{subfigure}{.43\columnwidth}
	\includegraphics[width=\columnwidth]{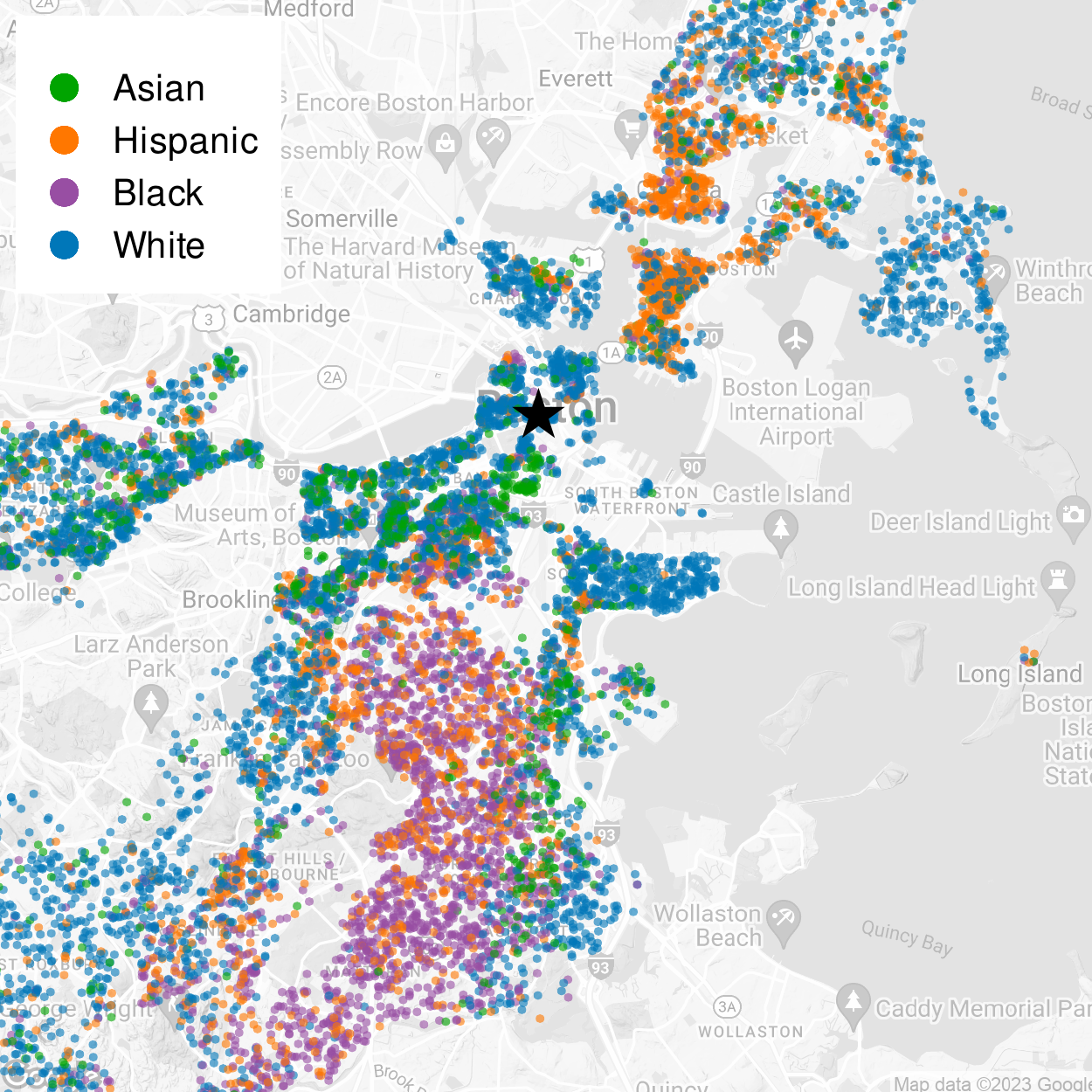}
		\caption{Suffolk County, MA residential distribution by race and ethnicity.}
		\label{fig:boston}
	\end{subfigure}
        \hspace{3mm}
	\begin{subfigure}{.46\columnwidth}
		\includegraphics[width=\columnwidth]{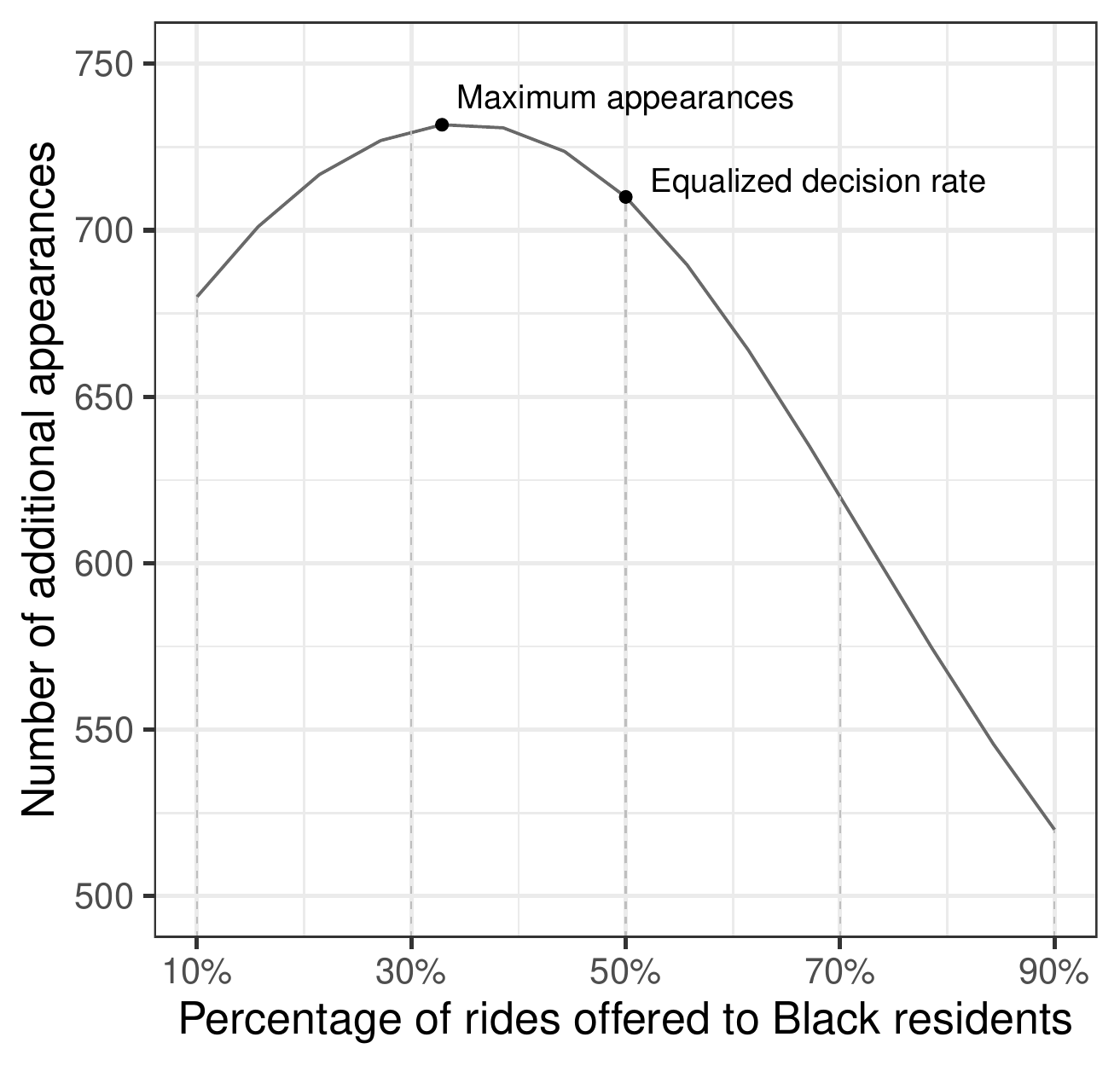}
		\caption{Trade-offs between maximizing court appearances and the demographic allocation of vouchers.}
		\label{fig:pareto-frontier}
	\end{subfigure}
	\vskip 0.1in
	\caption{The map in \subref{fig:boston} shows the geographic distribution of Suffolk County (Boston) residents, 
    with the star marking the location of the main county courthouse, where most individuals would be required to appear for court appointments. 
    In Suffolk County, White residents tend to live closer to the courthouse than Black residents.
    In \subref{fig:pareto-frontier}, we illustrate the range of possible policy options to provide a hypothetical population of residents with a free ride to court, where each option maximizes appearances for a given distribution of vouchers.
    For example,
    if one wants 50\% of vouchers to go to Black residents,
    given the existing budget, 710 additional people would get to court.
    Alternatively,
    under the same budget,
    a policy that aims to maximize appearances overall
    would allow 730 additional people to get to court
    and would have approximately 30\% of rides offered to Black residents.
    }
	\label{fig:map-and-pareto}
\end{figure}

Risk-based screening for diabetes is a setting where policy choices are not constrained by resources--- 
it is feasible to offer regular screening to every adult in the United States if that were determined to be medically advisable.
In other scenarios, however, policy decisions have to be made under significant resource constraints, leading to inherent trade-offs that complicate the design of equitable algorithms.
To illustrate, we transition 
from healthcare to the criminal-legal system,
and consider the problem of increasing court appearance rates among individuals with upcoming court dates.
In the United States, missed court dates typically prompt a judge to issue a ``bench warrant''---mandating the individual be arrested when they next encounter law enforcement---which in turn can lead to
days or even weeks in jail ~\citep{fishbane2020behavioral,mahoney2001pretrial}. 
Such incarceration imposes high costs on individuals and their communities, 
including job loss and social stigma~\citep{didwania2020immediate,dobbie2018,leslie2017,gupta2016,finlay2023measuring}.
Many people report missing their court appointments due to transportation barriers, and
so one promising proposal to improve appearance rates and reduce the resulting incarceration is to provide individuals with transportation vouchers (e.g., for public transit or ride-share services)~\citep{brough2022can,allen2022fta,L2BF}.
But these vouchers can be costly, and so such  programs may not be able to provide transportation assistance to every individual who might benefit from it.

Given the budget constraint, policymakers implementing transportation-assistance programs face a difficult trade-off: 
on one hand, they will want to spend their budget strategically, to increase appearances as much as possible (and, accordingly, maximally reduce incarceration);
on the other, they might also be interested in achieving a certain racial or ethnic balance among those who benefit from a voucher~\citep{L2BF}.
Consider the case of Boston, Massachusetts, where Black individuals tend to live farther away from the courthouse than White individuals, as shown in Figure~\ref{fig:boston}.
Because the costs of ride-share vouchers increase with the distance traveled, the demographic distribution of residents across the Boston area implies that, all else being equal, it would be more expensive to provide rides to Black individuals than to White individuals. 
As a result, a program solely focused on maximizing appearance rates would see more of its funds go to White clients. 
If instead one were to insist on equalized decision rates (i.e., offering transportation assistance to an equal proportion of White and Black individuals), this would necessarily mean that fewer appearances can be achieved. 

To make this trade-off more concrete, we describe the results of a simple, stylized simulation.
Suppose that 5,000 White and 5,000 Black individuals in a fictional city have upcoming court dates.
We imagine that it costs \$5 per mile to transport each individual from their home to the courthouse and back.
But, as in Boston, our hypothetical Black individuals on average live farther from the courthouse than our hypothetical White individuals. 
Finally, we suppose that for each individual $i$, they would successfully make it to court if provided a ride-share voucher but, if not provided a voucher, would appear with a known probability $p_i$---estimated, for example, with a model trained on historical court appearance data~\citep{L2BF}.
Assuming we have a transportation budget of \$10,000, 
Figure~\ref{fig:pareto-frontier}
shows how the number of additional court appearances varies with the demographic allocation of vouchers,
where each point on the curve corresponds to an allocation strategy that maximizes the number of appearances while ensuring a certain demographic composition of recipients.
Among these policy options, there is no one ``correct'' choice.
The \textit{best} choice  
will depend on one's preference for trading off the total number of court appearances with the distribution of vouchers across Black and White individuals, an idea we discuss more below.
For now, 
we note that 
certain formal fairness constraints---e.g., requiring an equal proportion of White and Black individuals receive vouchers---%
represent but one among several options to make that trade-off.

\section{A path forward}
\label{sec:thepath}

Our diabetes example suggests that adherence to popular fairness constraints often comes at the cost of inflicting additional burden on individuals, including those in marginalized groups.
In our ride-share example, where resources are limited, 
imposing formal fairness constraints can likewise result in allocation policies that do not reflect the preferences of stakeholders.   
This tension highlights the need for a robust discussion about the specific way in which proposed fairness constraints are connected to the inherently normative concept of equity.

On one hand, the constraints could be understood as an \textit{end} in and of itself. This conceptualization of fairness constraints is consistent with a deontological account of ethical decision-making, which postulates that an ethical decision is one that adheres to universally applicable, moral rules. Under this view, fairness criteria establish an outcome-independent set of constraints that should be imposed for their own sake, without regard to the specific results in a particular context.

A competing understanding of fairness constraints treats them not as an \textit{end}, but as a \textit{means} to achieve equitable, algorithmic decisions. Conceiving of constraints in this way is consistent with a \textit{consequentialist} account of ethical decision-making, whereby the morality of a decision is defined not by its dogmatic adherence to a set of rules, but by the outcomes it achieves. Under a consequentialist view, popular fairness constraints merely act as potentially useful heuristics to achieve desirable outcomes. However, if it can be demonstrated that a particular constraint imposes net burdens on marginalized groups, or society more generally, a consequentialist conception would counsel against dogmatic adherence to the constraint, as it would not achieve its desired goal of furthering equity in that case.

Although fairness constraints are frequently promoted and implemented, deeper discussions of their normative underpinnings are almost entirely absent from the literature (for a rare example, see~\citet{card2020consequentialism}). 
In principle, treating fairness constraints as a dogmatic principle or as a useful heuristic appear defensible. 
In our example of estimating diabetes risk, advocates in favor of race-blind tools may, for instance, argue that it is inherently wrong to make decisions for an individual based on their (immutable) group membership, or that race-based decision-making reinforces damaging beliefs about inherent differences between individuals of different racial groups. 
And perhaps some believe that these concerns outweigh the negative health  
effects that patients---particularly Asian patients---might experience under the race-blind algorithm.
But irrespective of one's particular preferences, we believe that a more considered discussion and explicit acknowledgement of the potential cost of these constraints is necessary to avoid inflicting accidental harms.

We now conclude our discussion by considering four technical and policy-related aspects of algorithm design that we believe are critical to building more equitable tools:
(1) grappling with the inherent trade-offs that we highlighted above; 
(2) checking the calibration of predictive models;
(3) judiciously selecting the target of prediction;
and (4) appropriately collecting training data.

\subsection*{Grappling with inherent trade-offs}

\begin{figure}[t]
	\centering
	\begin{subfigure}{.48\columnwidth}
	\includegraphics[width=\columnwidth]{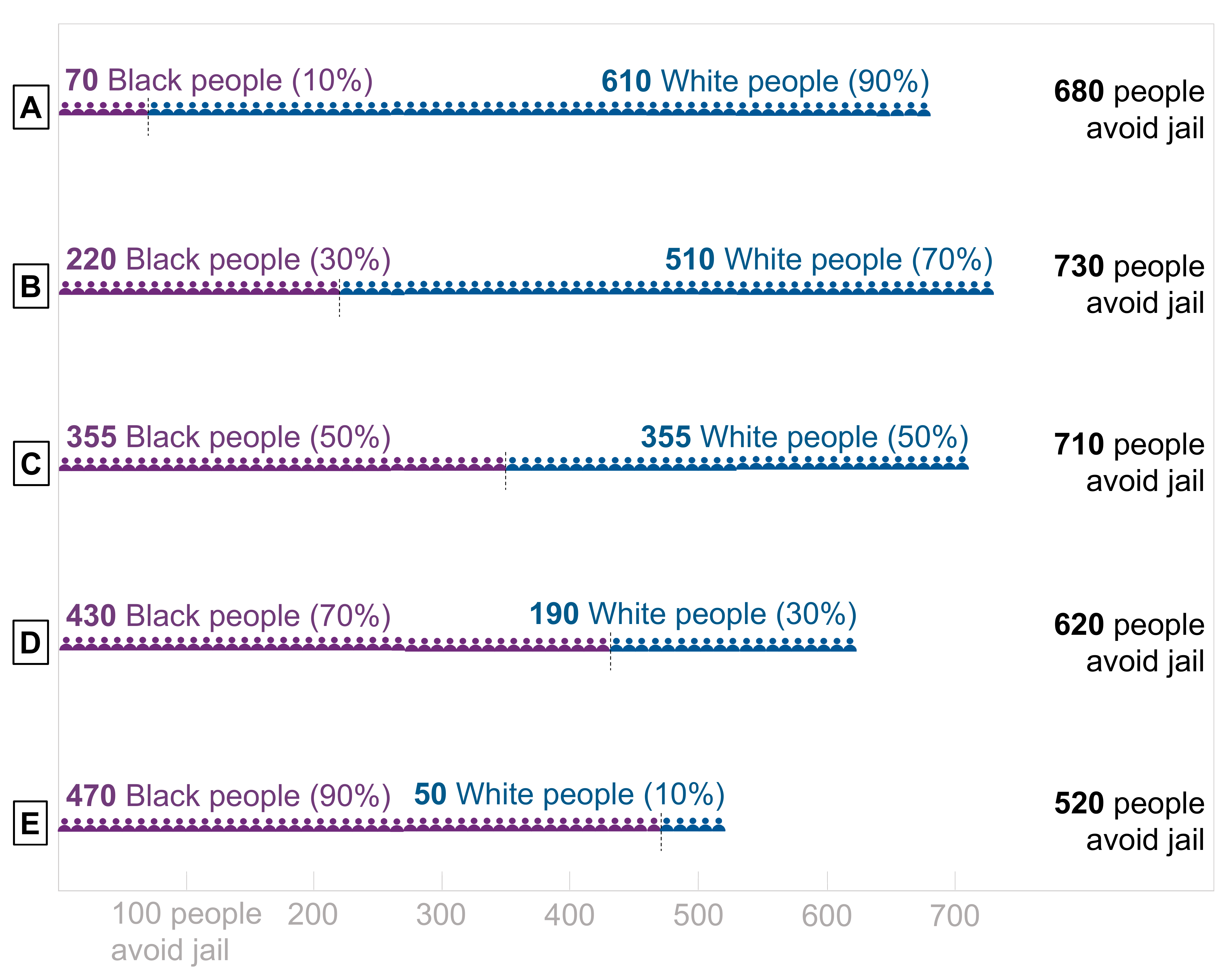}
		\caption{Options presented to survey participants, which involve a trade-off between minimizing overall incarceration and incarceration for Black people.}
		\label{fig:survey-graphic}
	\end{subfigure}
        \hspace{3mm}
	\begin{subfigure}{.48\columnwidth}
		\includegraphics[width=\columnwidth]{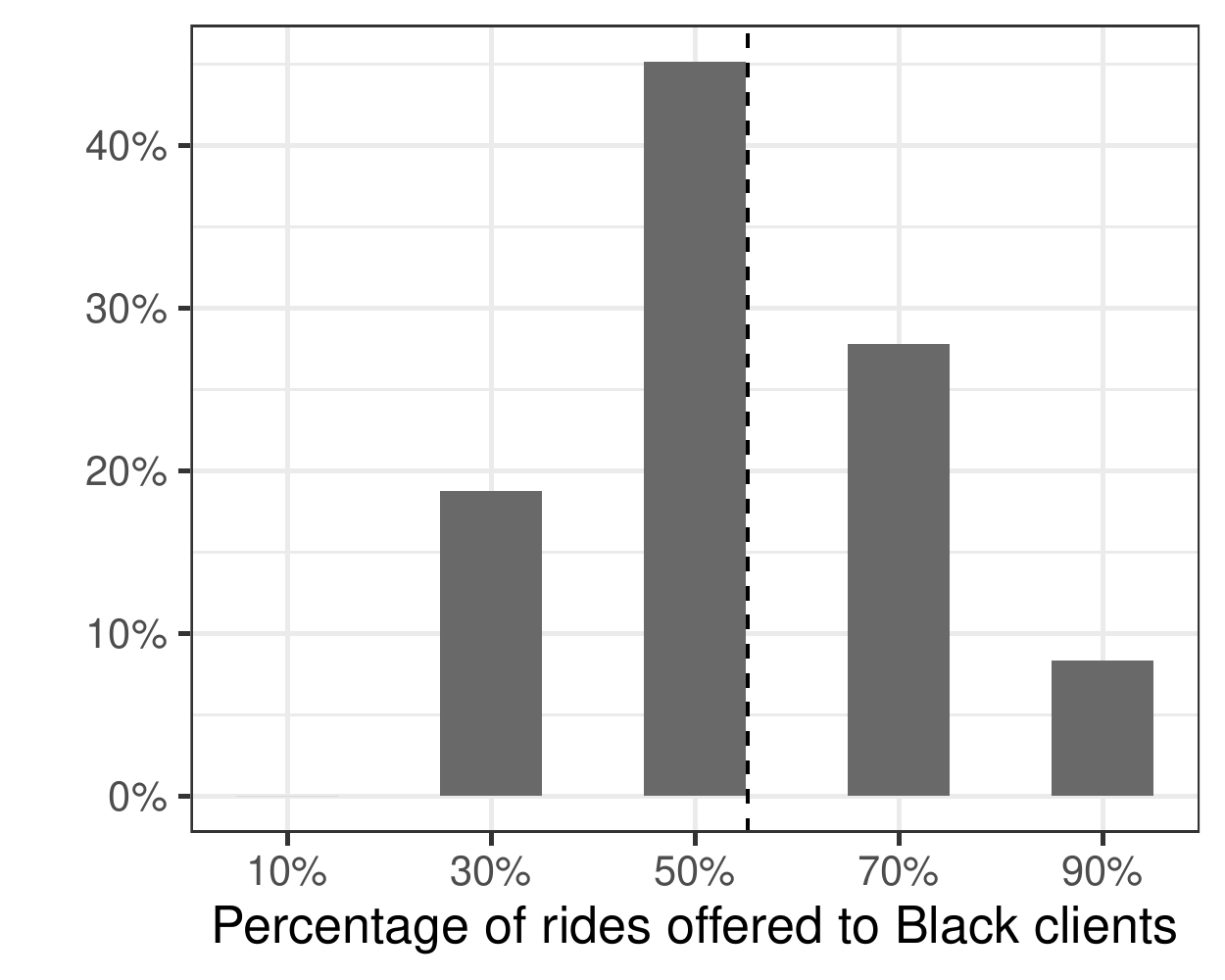}
		\caption{Survey responses corresponding to the options in Figure~\ref{fig:survey-graphic}, with the dashed vertical line at 55\% indicating the average response.}
		\label{fig:survey_preferences}
	\end{subfigure}
	\vskip 0.1in
	\caption{Preferences for allocating ride-share vouchers to a hypothetical population of residents depicted in Figure~\ref{fig:pareto-frontier}. In~\subref{fig:survey-graphic}, we show the options presented to survey respondents. Responses to this question are shown in \subref{fig:survey_preferences}.
    }
	\label{fig:survey_graphic_and_preferences}
\end{figure}

One way to make salient and arbitrate between the competing aspects of equity we describe above is to elicit the preferences of stakeholders.
To illustrate this approach in our example of allocating ride-share vouchers, we designed and administered a poll to a diverse sample of Americans~\citep{koenecke2023,L2BF}.\footnote{
We ran our survey on the Prolific platform.
By selecting the platform's ``representative sample'' option, the   distribution of self-identified sex, age, and ethnicity  in our sample matched the distribution in the U.S. Census. The survey resulted in 144 respondents.} Mirroring our simulation above, survey respondents were introduced to a hypothetical jurisdiction with an equal proportion of Black and White residents, with Black residents, on average, living farther from the courthouse than White residents. We then asked respondents to state how they would balance appearance rates (and, accordingly, incarceration for missed court appoitments) with the demographic distribution of transportation assistance. To aid in their decision, participants were shown the graphic depicted in Figure \ref{fig:survey-graphic}. 

The results of the survey are shown in Figure \ref{fig:survey_preferences}, and reveal that the respondents have highly heterogenous preferences. Most frequently, respondents prefer a policy that mirrors the demographics of the underlying population (Option C),
but many respondents prefer a different balance of ``efficiency" and ``demographic balance'', favoring policies that shift more resources towards Black individuals. Only 19\% of respondents prefer the efficiency-maximizing allocation policy that minimizes overall incarceration (Option B). 

Preferences elicited in this way are but one input into complex policy decisions. 
Further, while we surveyed a diverse sample of Americans, identifying the relevant stakeholders is itself a difficult problem, defying general prescriptions.
We hope, though, that this simple exercise demonstrates the feasibility of productively grappling with the thorny trade-offs at the heart of many policy design problems.

\subsection*{Checking calibration across groups}
As its name suggests, the primary objective of a risk assessment algorithm is to accurately estimate risk.
Without care, however, it is common for statistical algorithms to systematically overestimate risk for some groups and underestimate risk for others---a problem that is also referred to as ``miscalibration''.
For example, in Figure~\ref{fig:calibration-facet} (left), 
a 1\% estimated risk of diabetes corresponds to an observed diabetes rate of about 0.5\% for White individuals but about 1.6\% for Asian individuals.
These miscalibrated risk scores can lead to over-screening White patients and under-screening Asian patients, resulting in worse outcomes for individuals in both groups.
Similarly, gender-blind risk assessment tools commonly used in the criminal justice system to predict recidivism tend to systematically overestimate risk for women and underestimate risk for men~\citep{skeem2016gender}.
These miscalibrated estimates can in turn lead to incarcerating women who are much less likely to recidivate than their risk scores suggest.

Miscalibrated risk scores can typically be corrected by including group membership (e.g., race or gender) as a risk factor in the predictive model. 
Doing so, however, can run afoul of legal restrictions---for example, explicit considerations of race often receive heightened legal scrutiny in the United States, a standard that is notoriously difficult to satisfy. 
And even when legally permissible, race/ethnicity- and gender-aware algorithms may not be socially or politically acceptable.
As we discussed in the context of our running diabetes example, the use of race or ethnicity could reinforce pernicious attitudes about inherent differences between racial and ethnic groups, potentially outweighing the benefits of including these features in the models.
Regardless of whether one ultimately decides to include race, ethnicity, gender, or other sensitive attributes in risk assessment algorithms, we believe it is important to assess whether the risk estimates are calibrated (like we do in Figure~\ref{fig:calibration-facet}) in order to better understand the costs and benefits of algorithm design choices.

\subsection*{Selecting the target of prediction}

Even when including information about group membership, it is still possible to have miscalibrated estimates if the labels available in the data are an imperfect representation of the true target of the prediction. This occurrence is known as \textit{label bias}~\citep{label-bias,corbett2023measure}. For our running diabetes example, the NHANES data provides two pieces of information that we can use to construct a label for whether a patient has diabetes: (1) the results of a blood test administered to the entire survey population; and (2) whether the patient has ever received a diabetes diagnosis from a doctor. A diabetes label based solely on receiving a doctor's diagnosis is often tied to how regularly a patient is seen by a doctor. But one can imagine that the frequency of seeing a doctor varies considerably across a number of dimensions. 
For instance, given the same health-related attributes, Black patients tend to go to primary care physicians less often than White patients~\citep{arnett2016race}. 
And because the probability of detecting diabetes increases with the frequency with which a patient goes to the doctor, a dataset that records the presence of diabetes using a doctor's assessment might systematically under-record the presence of diabetes in racial minorities. Consequently, a model trained on such a dataset would systematically underestimate the \emph{true} diabetes risk of racial minorities, even if it is well calibrated to predicting a doctor's diabetes diagnosis.

\begin{figure}[t]
\centering
\includegraphics[width=0.5\columnwidth]{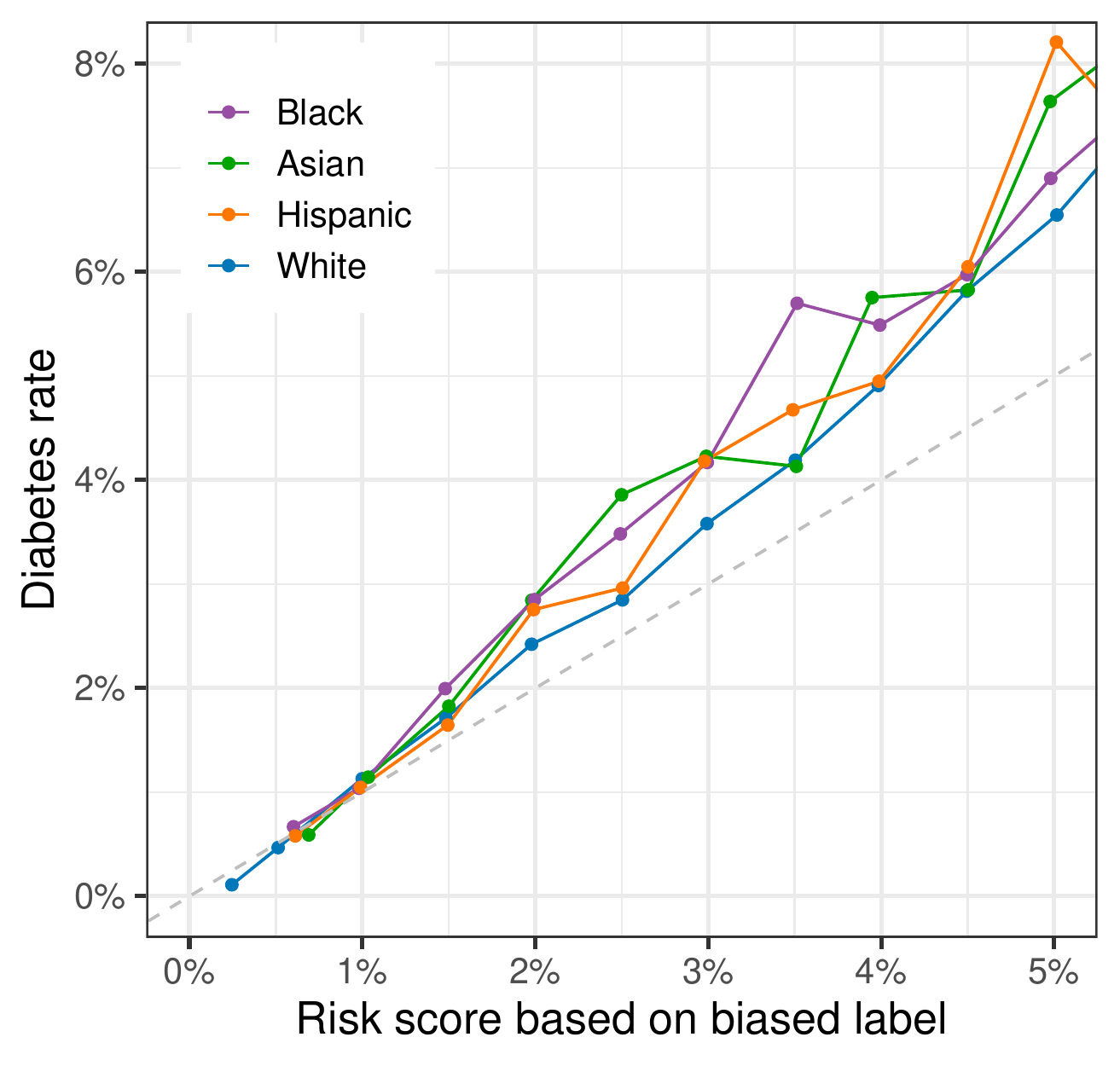}
\caption{ In contrast to the risk scores presented in Figure \ref{fig:calibration-facet} which predict the prevalence of diabetes from blood tests and doctors' diabetes diagnoses, here the models are only trained to predict a doctor's diabetes diagnosis. The model inputs are age, race/ethnicity and BMI. Likely due in part to racial and ethnic disparities in healthcare access, predicting a doctor's diabetes diagnosis introduces bias into the model when compared against the results from the combined label. We observe that Asian, Black, and Hispanic Americans have higher true diabetes risk than White Americans with the same nominal risk under the model.}
\label{fig:label-bias-calibration}
\end{figure}

Figure \ref{fig:label-bias-calibration} lends empirical support to these theoretical concerns by illustrating the negative effects of using a doctor's diagnosis as a proxy for the presence of diabetes. It shows the risk scores of a model trained to predict the proxy (doctor's diabetes diagnosis) using a patient's age, race/ethnicity, and BMI. Yet, in spite of supplying the model with data on group membership---race and ethnicity, in this case---the resulting risk scores are still miscalibrated when compared against the true label (blood test \textit{or} doctor diagnosis) for each group. The risk scores produced using the proxy underestimate risk for all groups, but miscalibration is especially pronounced for Black, Asian, and Hispanic patients. This pattern may in part stem from known racial disparities in healthcare access, with patients from minority racial groups less likely to have received a diabetes diagnosis from their doctor. The miscalibrated risk scores could lead to under-screening of patients from \textit{all} groups, but most severely for racial and ethnic minorities.

Unfortunately, one's ability to mitigate the effects of label bias is often heavily constrained by the data collection process and availability. 
One possible solution is to adjust the target of prediction by focusing on or leveraging other outcomes that are less likely to exhibit bias. In our running diabetes example, we accomplish this by constructing a diabetes label using \textit{both} the blood test results provided in the survey \textit{and} whether the patient had ever received a diabetes diagnosis from a doctor. Because the blood test was administered to the entire population, using this information to construct the diabetes label fills in informational gaps that arise from relying solely on past diabetes diagnoses from a doctor. Countering label bias is challenging and there is rarely a perfect solution, but when designing algorithmic tools for decision-making, it is imperative to scrutinize the data variables used--- especially that of the target of interest---for potential sources of bias and mitigate downstream effects to the greatest extent possible.

\subsection*{Collecting training data}
Finally, we consider the role of training data in equitable algorithm design.
As a general heuristic, we advise to train algorithms on data that are representative of the population to which the algorithms will be applied. Failure to do so can lead to starkly inequitable outcomes. 
For example, in an analysis of automated speech recognition systems, 
\citet{koenecke2020racial} found that state-of-the-art models made twice as many errors transcribing Black speakers than they did for White speakers, a disparity that likely stemmed from a relative sparsity of speech data from Black speakers used to train the models.
Similarly, in an analysis of image analysis tools, \citet{buolamwini2018gender} found that
popular algorithms performed significantly worse at classifying the gender of dark-skinned individuals compared to light-skinned individuals, likely due to a lack of dark-skinned faces in the training data---and the performance was worst for dark-skinned women.\footnote{%
In both of these examples, error rates by race and skin tone are a useful metric for auditing the algorithms because one would not generally expect the difficulty of transcribing speech or identifying gender to vary substantially across these categories. 
In contrast, in many risk assessment settings like our diabetes example, we expect risk distributions to differ across racial/ethnic and gender groups, limiting the diagnostic value of comparing error rates.
}

Training algorithms with representative data is a useful starting point, but, like in many other contexts we discuss here, there are subtleties to consider.
For example, given a limited budget,
the optimal data collection strategy depends on the statistical structure of the underlying population, the cost of collecting data from different subgroups, and the relative value of model performance across subgroups~\citep{cai2022adaptive}.
In particular, if the connection between features and outcomes is similar across subgroups, one might trade representativeness for more data from the subgroups for whom data acquisition is less costly.
Conversely, if certain groups have idiosyncratic statistical properties, one might choose to oversample from them.
In short, and in line with our general philosophy, it is important to carefully consider the trade-offs inherent to different data collection strategies.

\section{Conclusion}
With the proliferation of algorithmically guided decision-making in healthcare, the criminal-legal system, banking, and beyond, there is increasing need to ensure that algorithms are fair. 
A plethora of formal fairness metrics and design principles have been proposed in recent years, particularly in the computer science community.
But, as we have argued here, popular approaches to fairness often lead to worse outcomes for individuals, including those from marginalized communities.
In some cases, the conflict between formal fairness constraints and equitable outcomes suggests shortcomings of the constraints themselves.
For instance, in our running diabetes example, 
it seems difficult to justify equalizing error rates on consequentialist grounds.
In other cases, though, the tension is harder to allay. 
Diabetes risk estimates that consider race and ethnicity may lead to more accurate screening decisions, but such non-blind algorithms might also validate and encourage insidious beliefs in inherent differences across groups, with possible negative repercussions for marginalized patients, many of whom already have significant distrust in the U.S.\ healthcare system~\citep{boulware2003race,arnett2016race}.
There are often no easy answers to these difficult trade-offs, but we hope our discussion equips researchers, practitioners, and policymakers to make more informed choices.

\section*{Acknowledgements}
We thank Sam Corbett-Davies, Johann Gaebler, Avi Feller, David Kent, Keren Ladin, Hamed Nilforoshan, and Ravi Shroff for helpful conversations. 
We draw in this paper from 
a more technical exposition of algorithmic fairness by \citet{corbett2023measure}.
Our work was supported by grants from the Harvard Data Science Institute, the Stanford Impact Labs, and Stanford Law School.
Code to reproduce our analysis is available at: https://github.com/madisoncoots/equitable-algorithms.

\clearpage
\bibliography{refs}

\end{document}